\documentclass[conference]{IEEEtran}
\IEEEoverridecommandlockouts

\usepackage{cite}
\usepackage{amsmath,amssymb,amsfonts}
\usepackage{algorithmic}
\usepackage{graphicx}
\usepackage{textcomp}
\usepackage{xcolor}
\def\BibTeX{{\rm B\kern-.05em{\sc i\kern-.025em b}\kern-.08em
    T\kern-.1667em\lower.7ex\hbox{E}\kern-.125emX}}
\begin{document}

\title{AIMS: An Adaptive Integration of Multi-Sensor Measurements for Quadrupedal Robot Localization\\
\thanks{ This work was supported in part by the National Natural Science Foundation of China under Grant 62573078, the Scientific and Technological Research Program of Chongqing Municipal Education Commission under Grant
	No.KJZD-K202200603, the Natural Science Foundation of Chongqing under
	Grant No.CSTB2024TIAD-STX0023 and No.CSTB2023TIAD-STX0016, 
}
\thanks{
	*Corresponding author. Email: haozhu1982@gmail.com.}
}

\author{
	\IEEEauthorblockN{1\textsuperscript{st} Yujian Qiu$^*$}
	\IEEEauthorblockA{\textit{College of Automation} \\
		\textit{Chongqing University of Posts and Telecommunications}\\
		Chongqing, China \\
		qiuyujiancqupt@163.com} \\
	\IEEEauthorblockN{3\textsuperscript{rd} Wen Yang}
	\IEEEauthorblockA{\textit{College of Automation} \\
		\textit{Chongqing University of Posts and Telecommunications}\\
		Chongqing, China \\
		S240301025@stu.cqupt.edu.cn} \\
	\and
	\IEEEauthorblockN{2\textsuperscript{nd} Yuqiu Mu}
	\IEEEauthorblockA{\textit{College of Automation} \\
		\textit{Chongqing University of Posts and Telecommunications}\\
		Chongqing, China \\
		S230301037@stu.cqupt.edu.cn} \\
	\IEEEauthorblockN{4\textsuperscript{th} Hao Zhu}
	\IEEEauthorblockA{\textit{College of Automation} \\
		\textit{Chongqing University of Posts and Telecommunications}\\
		Chongqing, China \\
		zhuhao@cqupt.edu.cn} \\
}
\maketitle

\begin{abstract}
This paper addresses the problem of accurate localization for quadrupedal robots operating in narrow tunnel-like environments. Due to the long and homogeneous characteristics of such scenarios, LiDAR measurements often provide weak geometric constraints, making traditional sensor fusion methods susceptible to accumulated motion estimation errors. To address these challenges, we propose AIMS, an adaptive LiDAR-IMU-leg odometry fusion method for robust quadrupedal robot localization in degenerate environments. The proposed method is formulated within an error-state Kalman filtering framework, where LiDAR and leg odometry measurements are integrated with IMU-based state prediction, and measurement noise covariance matrices are adaptively adjusted based on online degeneracy-aware reliability assessment. Experimental results obtained in narrow corridor environments demonstrate that the proposed method improves localization accuracy and robustness compared with state-of-the-art approaches.
\end{abstract}


\section{Introduction}
Safety inspection is an essential routine task in industrial facilities. Traditional manual inspection is often inefficient, repetitive, and potentially dangerous, motivating the increasing deployment of intelligent mobile robots for autonomous inspection tasks\cite{gao2018autonomous,jia2022intelligent}. In such applications, accurate and continuous localization is a fundamental prerequisite for reliable autonomous navigation and inspection performance\cite{li2023multi,zhu2017overview,xue2025vgpnet}.

Simultaneous localization and mapping (SLAM) has been widely recognized as a core technology for autonomous robots\cite{losch2018design,zhu2021novel}, enabling online estimation of robot states without requiring prior environmental information\cite{zou2021comparative}. Existing SLAM approaches can be broadly categorized into vision-based and LiDAR-based methods. Vision-based SLAM systems have demonstrated impressive performance in feature-rich and well-lit environments, leveraging visual features and visual-inertial fusion for state estimation\cite{qin2018vins,campos2021orb,xu2025airslam}. However, in narrow inspection areas such as long corridors or confined passages, illumination conditions are often insufficient\cite{xue2025feature}, which significantly degrades the reliability of visual perception and limits the applicability of vision-based SLAM methods\cite{tardioli2019ground}.

Compared with visual sensors, three-dimensional LiDAR provides a wide field of view, long sensing range, and illumination-invariant perception\cite{guo2022lirtest}, making LiDAR-based SLAM a more suitable choice for inspection robots operating in confined environments\cite{cadena2017past}. As a result, LiDAR-based and LiDAR-IMU fusion methods have become the dominant solutions for robot localization in such scenarios\cite{zhou2021t}. Nevertheless, in long and narrow corridors with homogeneous structures, the geometric constraints provided by LiDAR measurements become weak, leading to severe perceptual degradation. Under these conditions, the robot motion is often underestimated, and the reconstructed map tends to be shorter than the actual environment scale, resulting in accumulated localization drift and inconsistency\cite{zhu2022variational,zhu2024adaptive}.

To address this issue, In \cite{zhu2021adaptive}, a novel variational Bayesian (VB) adaptive Kalman filter with inaccurate nominal process and measurement noise covariances (PMNC) in the presence of outliers is proposed. The LINS\cite{qin2020lins} proposes a tightly-coupled fusion scheme of a 6-axis IMU and a 3D LiDAR, enabling robots to achieve robust and efficient navigation in feature-less environments. The FAST-LIO2\cite{xu2022fast} directly registers raw point clouds to the map without explicit feature extraction, which enables the exploitation of subtle geometric cues in the environment and helps mitigate perceptual degeneration. The EKF-LOAM\cite{junior2022ekf} presents a simple and lightweight adaptive covariance matrix based on the number of LiDAR detected geometric features. The covariance matrix is used to adaptively fuse LiDAR measurements. In \cite{xu2021indoor}, an adaptive federated Kalman filter (AFKF) algorithm is proposed. According to the fault detector is added to detect extreme abnormal condition. In \cite{han2023dams}, a light-weight iEKF-based LiDAR-inertial odometry system is presented, which utilizes a degeneration-aware and modular sensor-fusion pipeline that takes both LiDAR points and relative pose from another odometry as the measurement in the update process only when degeneration is detected. While these methods demonstrate improved performance in environments containing intermittent feature-rich regions, they generally assume the existence of non-degenerate areas along the robot trajectory. If perceptual degradation persists over extended distance, these assumptions may no longer hold, and localization failures can still occur.

Motivated by these observations, this paper proposes an adaptive LiDAR-IMU-leg odometry fusion method for robust localization of quadrupedal robots in narrow corridor environments. The proposed method is formulated within an error-state Kalman filtering framework, where LiDAR and leg odometry measurements are jointly integrated with IMU-based state prediction. A degeneracy-aware reliability assessment mechanism is introduced to continuously evaluate perceptual degradation, and the resulting reliability indicators are used to adaptively regulate the information contribution of different sensors during fusion. Experimental results obtained using a quadrupedal robot in narrow corridor environments demonstrate that the proposed method achieves improved localization accuracy and robustness compared with state-of-the-art approaches.

\section{The Proposed Method}
\label{sec:method}

\subsection{System Overview}

\begin{figure}[htbp]
	\centering
	\includegraphics[width=\linewidth]{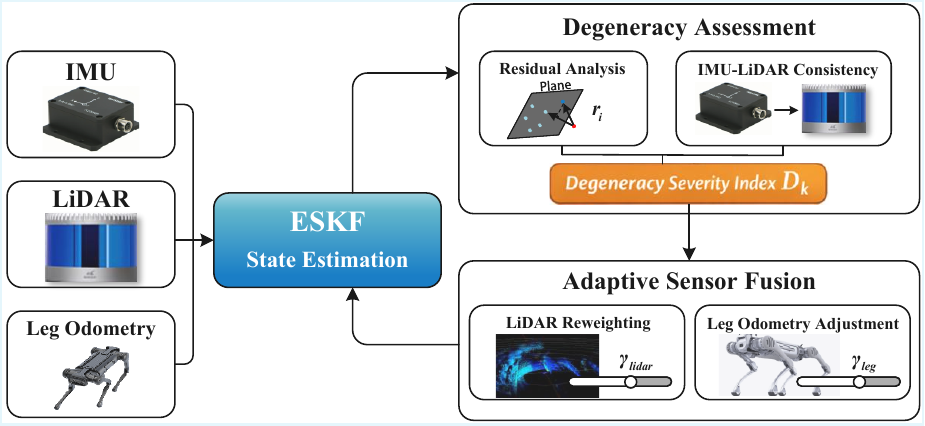}
	\caption{Framework of the proposed AIMS.}
	\label{fig:overview}
\end{figure}

The proposed AIMS framework aims to achieve robust localization for quadrupedal robots operating in long and narrow inspection environments. 
It adopts a tightly-coupled multisensor fusion architecture based on an error-state Kalman filter (ESKF).
IMU measurements are used for continuous state prediction, while LiDAR and leg odometry measurements are incorporated as motion constraints during the correction stage.
The overall framework is designed to explicitly handle perceptual degeneration by online reliability assessment and adaptive sensor reweighting, as shown in Fig.~\ref{fig:overview}.

\subsection{IMU-Based State Prediction}

The system state $\mathbf{x}$ is defined as
\begin{equation}
	\mathbf{x} =
	\left[
	\mathbf{R},
	\mathbf{p},
	\mathbf{v},
	\mathbf{b}_g,
	\mathbf{b}_a
	\right]
\end{equation}
where $\mathbf{R} \in SO(3)$ denotes orientation, $\mathbf{p} \in \mathbb{R}^3$ position, $\mathbf{v} \in \mathbb{R}^3$ velocity, and $\mathbf{b}_g$, $\mathbf{b}_a$ the gyroscope and accelerometer biases.

Given IMU angular velocity $\boldsymbol{\omega}_m$ and linear acceleration $\mathbf{a}_m$, the state is propagated using the standard inertial motion model.
The predicted state and covariance are denoted as $\hat{\mathbf{x}}_{k|k-1}$ and $\mathbf{P}_{k|k-1}$, respectively.

\subsection{LiDAR and Leg Odometry Measurement Models}

LiDAR measurements are modeled using point-to-plane constraints constructed from a local map.
For each LiDAR point $\mathbf{p}_i$, the residual is defined as
\begin{equation}
	r_i =
	\mathbf{n}_i^\top
	\left(
	\mathbf{R}\mathbf{p}_i + \mathbf{t} - \mathbf{q}_i
	\right)
\end{equation}
where $\mathbf{q}_i$ and $\mathbf{n}_i$ denote the corresponding plane point and normal vector.

Leg odometry provides relative motion constraints between consecutive states.

\subsection{Degeneracy-Aware Reliability Assessment}

In long and homogeneous environments, perceptual degeneration is a continuous process that evolves over time.
To characterize the severity of degeneration and its impact on state estimation, the proposed method performs a continuous reliability assessment based on LiDAR residual statistics and IMU-LiDAR consistency.

As defined in Sec.~C, LiDAR measurements generate a set of point-to-plane residuals $\{r_i\}_{i=1}^{N}$.
Beyond their magnitude, the distribution of residuals reflects the effective observability of the underlying state.

Let $\bar{r}$ and $\sigma_r^2$ denote the mean and variance of residuals:
\begin{equation}
	\bar{r} = \frac{1}{N} \sum_{i=1}^{N} r_i, \quad
	\sigma_r^2 = \frac{1}{N} \sum_{i=1}^{N} (r_i - \bar{r})^2
\end{equation}

In perceptually degenerate scenarios, residuals tend to concentrate along a limited number of directions, resulting in low variance despite small residual magnitudes.
To quantify this effect, an effective observability metric is defined as
\begin{equation}
	\mathcal{O}_{\text{lidar}} =
	\frac{\sigma_r^2}{\sigma_r^2 + \sigma_0^2}
\end{equation}
where $\sigma_0^2$ is a normalization constant reflecting nominal residual dispersion. A smaller $\mathcal{O}_{\text{lidar}}$ indicates reduced excitation of the measurement model and limited geometric observability.

Residual-based observability alone is insufficient to characterize degeneration, as small residuals may also arise from accurate alignment.
Therefore, the consistency between LiDAR updates and IMU prediction is further examined.

Let the innovation vector be defined as
\begin{equation}
	\mathbf{y}_k =
	\hat{\mathbf{x}}_{k}^{\text{lidar}} \ominus \hat{\mathbf{x}}_{k|k-1}
\end{equation}
where $\ominus$ denotes the difference operation on the state manifold, $\hat{\mathbf{x}}_{k|k-1}$ is the IMU-predicted state and $\hat{\mathbf{x}}_{k}^{\text{lidar}}$ is the LiDAR-updated state.

The normalized squared is computed as
\begin{equation}
	\mathcal{C}_{\text{IL}} =
	\mathbf{y}_k^\top
	\mathbf{P}_{k|k-1}^{-1}
	\mathbf{y}_k
\end{equation}
where $\mathbf{P}_{k|k-1}$ denotes the predicted error covariance.

To jointly characterize perceptual degeneration, a Degeneracy Severity Index is defined as
\begin{equation}
	\mathcal{D}_k =
	w_1 (1 - \mathcal{O}_{\text{lidar}})
	+ w_2 \frac{\mathcal{C}_{\text{IL}}}{\mathcal{C}_{\text{IL}} + \kappa}
\end{equation}
where $w_1$ and $w_2$ are weighting factors satisfying $w_1 + w_2 = 1$, and $\kappa$ is a saturation constant.

The index $\mathcal{D}_k \in [0,1]$ reflects the severity of perceptual degeneration, with larger values indicating more severe degeneration.

\subsection{Adaptive Multisensor Fusion}

The estimated degeneracy severity $\mathcal{D}_k$ is directly incorporated into the multisensor fusion process to regulate information injection from different sensors. Instead of directly scaling measurement noise, the proposed method modulates the effective information contribution of LiDAR measurements.
The LiDAR reliability factor is defined as
\begin{equation}
	\gamma_{\text{lidar}} =
	\exp(-\eta \mathcal{D}_k)
\end{equation}
where $\eta$ controls the sensitivity to degeneration.

Accordingly, the LiDAR measurement covariance is adjusted as
\begin{equation}
	\mathbf{R}_{\text{lidar}}' =
	\frac{1}{\gamma_{\text{lidar}}}
	\mathbf{R}_{\text{lidar}}
\end{equation}

Leg odometry provides complementary proprioceptive motion constraints that are less affected by environmental structure.
However, excessive reliance on leg odometry may lead to drift accumulation.

To achieve balanced fusion, the leg odometry reliability is coupled with LiDAR reliability:
\begin{equation}
	\gamma_{\text{leg}} =
	\gamma_{\min}
	+ (1 - \gamma_{\min})(1 - \mathcal{D}_k)
\end{equation}
where $\gamma_{\min}$ ensures a minimum confidence level.

The leg odometry covariance is adjusted as
\begin{equation}
	\mathbf{R}_{\text{leg}}' =
	\frac{1}{\gamma_{\text{leg}}}
	\mathbf{R}_{\text{leg}}
\end{equation}

To avoid abrupt changes in sensor weighting, the degeneracy severity index is temporally smoothed:
\begin{equation}
	\tilde{\mathcal{D}}_k =
	\alpha \tilde{\mathcal{D}}_{k-1}
	+ (1 - \alpha) \mathcal{D}_k
\end{equation}
where $\alpha \in [0,1)$ controls the smoothing factor.

The smoothed index $\tilde{\mathcal{D}}_k$ is used in all adaptive reweighting operations, ensuring stable and consistent state estimation.

\section{Experiments}

In this section, a series of experiments are conducted to evaluate the effectiveness and robustness of the proposed method for quadrupedal robot localization in narrow environments.

\subsection{Experimental Platform and Evaluation Metrics}
\begin{figure}[htbp]
	\centering
	\includegraphics[width=0.8\linewidth]{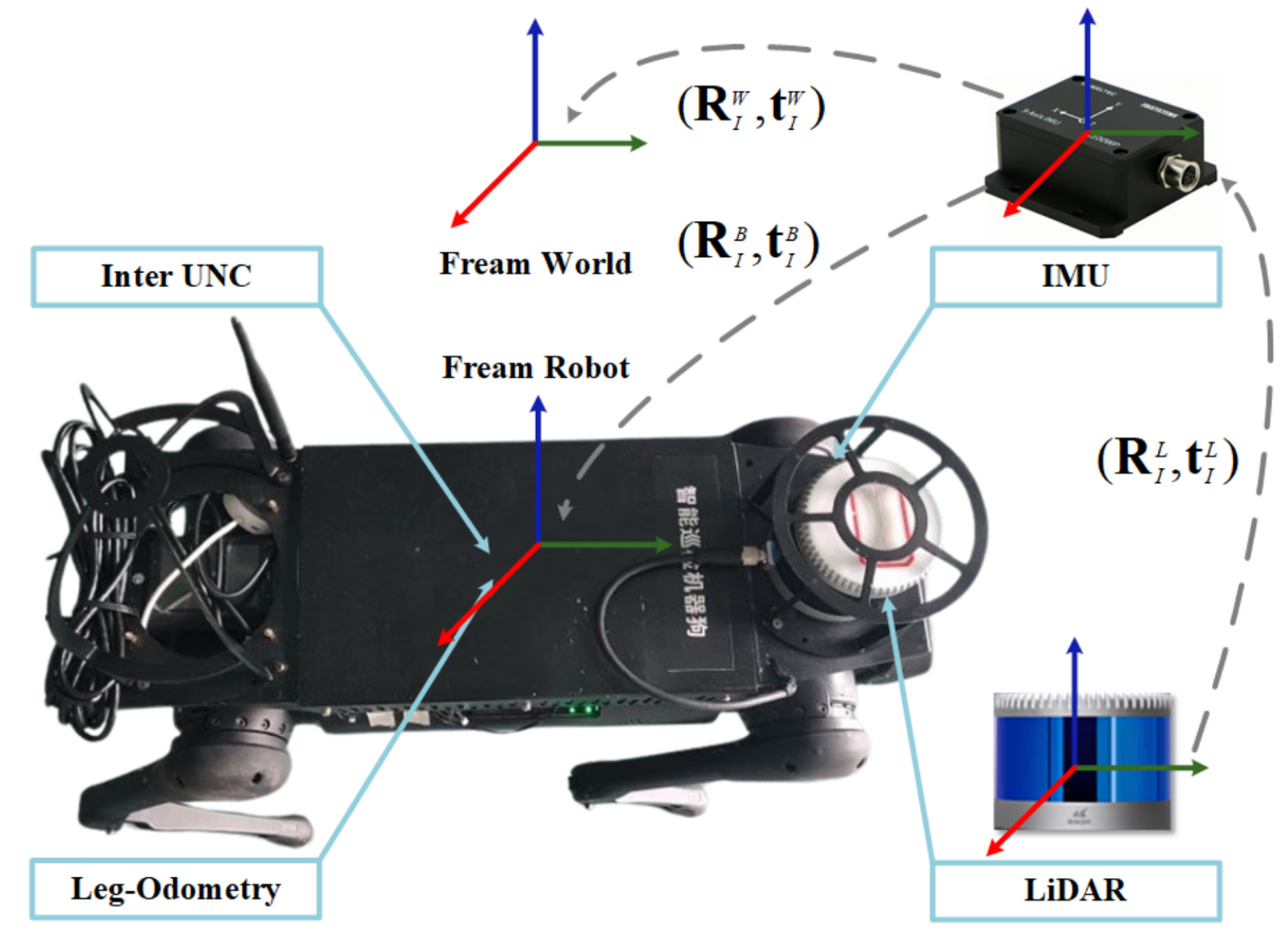}
	\caption{Diagram of Coordinate Frames.}
	\label{fig:dog}
\end{figure}
All experiments are conducted using a quadrupedal robot platform, as shown in Fig.~\ref{fig:dog}. The robot is equipped with a 3D LiDAR operating at 10~Hz, an IMU with a sampling rate of 50~Hz, and a leg odometry module providing measurements at 200~Hz. All sensors are rigidly mounted on the rear part of the robot body. An onboard Intel NUC8i5BEH computer is used to process all sensor data in real time.

Since ground-truth poses are not available in indoor environments, two evaluation metrics are adopted in this work: the degeneration error and the endpoint drift error.

The degeneration error is defined to evaluate the consistency between the estimated trajectory and the actual traveled distance:
\begin{equation}
	M_{\text{list}} = \frac{\sum_{i=1}^{N} \left| \hat{d}_i - d_i \right|}{\sum_{i=1}^{N} d_i}
\end{equation}
where $\hat{d}_i$ denotes the estimated traveled distance of the $i$-th segment, $d_i$ is the corresponding manually measured ground-truth distance, and $N$ is the total number of motion segments. When the robot moves approximately along a straight line, distances are measured using a tape measure. A smaller value of $M_{\text{list}}$ indicates higher localization accuracy, while values greater than 1 imply severe system drift.
The endpoint drift error is used to evaluate the accumulated drift when the robot returns to its starting position:
\begin{equation}
	M_{\text{end}} = \left\| \hat{\mathbf{p}}_{\text{end}} - \mathbf{p}_{\text{start}} \right\|
\end{equation}
where $\hat{\mathbf{p}}_{\text{end}}$ and $\mathbf{p}_{\text{start}}$ represent the estimated final position and the initial position of the robot, respectively.

\subsection{Indoor Long Corridor Experiment}

The indoor long corridor experiment is conducted in an automotive electronics laboratory at Chongqing University of Posts and Telecommunications. The laboratory consists of a narrow corridor with limited geometric structure, where LiDAR perception is prone to degeneration.

\begin{figure}[htbp]
	\centering
	\includegraphics[width=\linewidth]{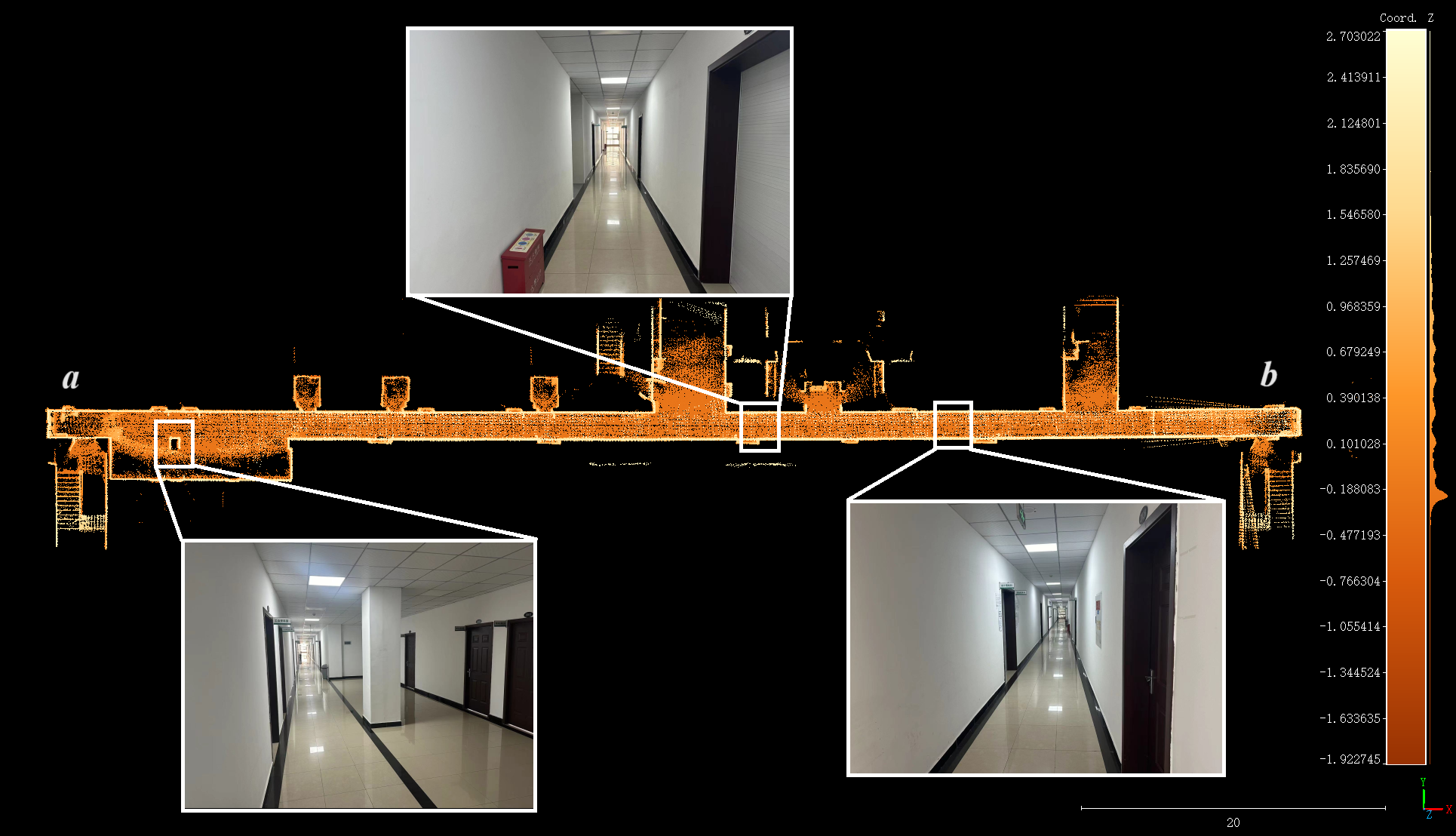}
	\caption{The 3D point cloud map constructed by the proposed method in the long corridor experiment.}
	\label{fig:proposed_method_map}
\end{figure}

Fig.~\ref{fig:proposed_method_map} illustrates the 3D point cloud map constructed by the proposed method. The embedded images show the corresponding real-world scenes. The color bar on the right indicates the height distribution of the point cloud. The robot moves from point $a$ to point $b$ and then returns to $a$.

\begin{figure}[htbp]
	\centering
	\includegraphics[width=\linewidth]{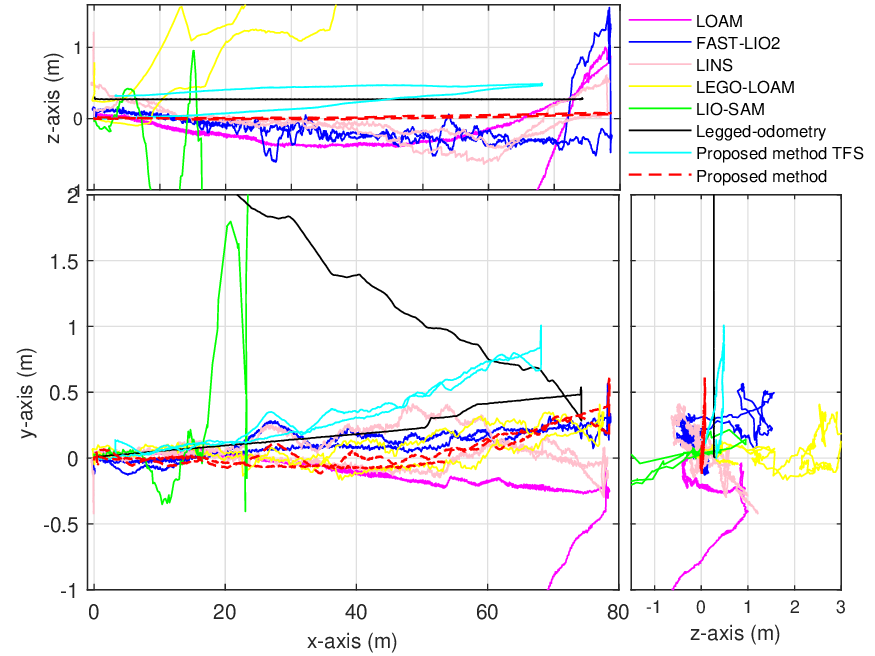}
	\caption{Trajectories estimated by legged odometry and other methods in the long corridor experiment.}
	\label{fig:all_traj}
\end{figure}

Five state-of-the-art SLAM methods are selected for comparison: LOAM\cite{zhang2014loam}, LEGO-LOAM\cite{shan2018lego}, FAST-LIO2\cite{xu2022fast}, LINS\cite{qin2020lins}, and LIO-SAM\cite{shan2020lio}. In addition, an ablation study is performed to analyze the influence of the adaptive fusion strategy. For this purpose, a variant of the proposed method with all adaptive modules disabled is implemented, referred to as \textit{Proposed method TFS}. Specifically, the adaptive weighting modules for both LiDAR–IMU measurements and leg odometry measurements are disabled.
The estimated trajectories of all methods are shown in Fig.~\ref{fig:all_traj}.

From the trajectory comparison, it can be observed that leg odometry produces erroneous measurements when the robot turns around at point $b$. Starting from the origin, FAST-LIO2 and the proposed method follow similar trajectories until the robot reaches approximately $(72\,\text{m}, 0.25\,\text{m})$. However, FAST-LIO2 exhibits an upward motion due to insufficient ground constraints, which can be confirmed by the estimated $z$-axis component. LIO-SAM requires higher-frequency IMU measurements, while the IMU onboard the quadrupedal robot operates at only 50~Hz. Combined with sparse LiDAR points in the corridor, LIO-SAM drifts after approximately 10 meters. LINS shows excellent performance along the $z$-axis and minimal degeneration along the $x$-axis, but noticeable fluctuations are observed along the $y$-axis.

The straight-line distance between points $a$ and $b$ is manually measured as 78.60~m. Since the robot returns to its starting position, both degeneration error and endpoint drift error are computed. Table~\ref{tab:corridor} reports the quantitative results, where $M_{\text{list}}(a\!-\!b)$ and $M_{\text{list}}(b\!-\!a)$ denote the degeneration degree for the forward and return paths, respectively.

\begin{table}[htbp]
	\centering
	\caption{Quantitative Comparison on the Corridor Dataset}
	\label{tab:corridor}
	\begin{tabular}{lcccc}
		\hline
		Method & $M_{\text{list}}(a\!-\!b)$ & $M_{\text{list}}(b\!-\!a)$ & $M_{\text{end}}$ \\
		\hline
		\multicolumn{4}{c}{\footnotesize{with "-" and bold number indicates meaningless and best result}} \\
		\hline
		LOAM & 0.15 & 1.73 & 16.32 \\
		FAST-LIO2 & 0.12 & 0.18 & 0.36 \\
		LINS & 0.14 & 0.16 & 1.29 \\
		LEGO-LOAM & 0.89 & 0.92 & 0.79 \\
		LIO-SAM & - & - & - \\
		Legged Odometry & 6.16 & 15.97 & 7.01 \\
		Proposed method TFS & 14.83 & 17.04 & 3.44 \\
		Proposed method & \textbf{0.07} & \textbf{0.08} & \textbf{0.16} \\
		\hline
	\end{tabular}
\end{table}

The proposed method achieves the best performance in both metrics. Proposed method TFS suffers from larger degeneration due to indiscriminate fusion of erroneous measurements. FAST-LIO2, LINS, and LEGO-LOAM show similar performance to the proposed method, and their behaviors are further analyzed through mapping results.

\begin{figure*}[t]
	\centering
	\includegraphics[width=0.95\textwidth]{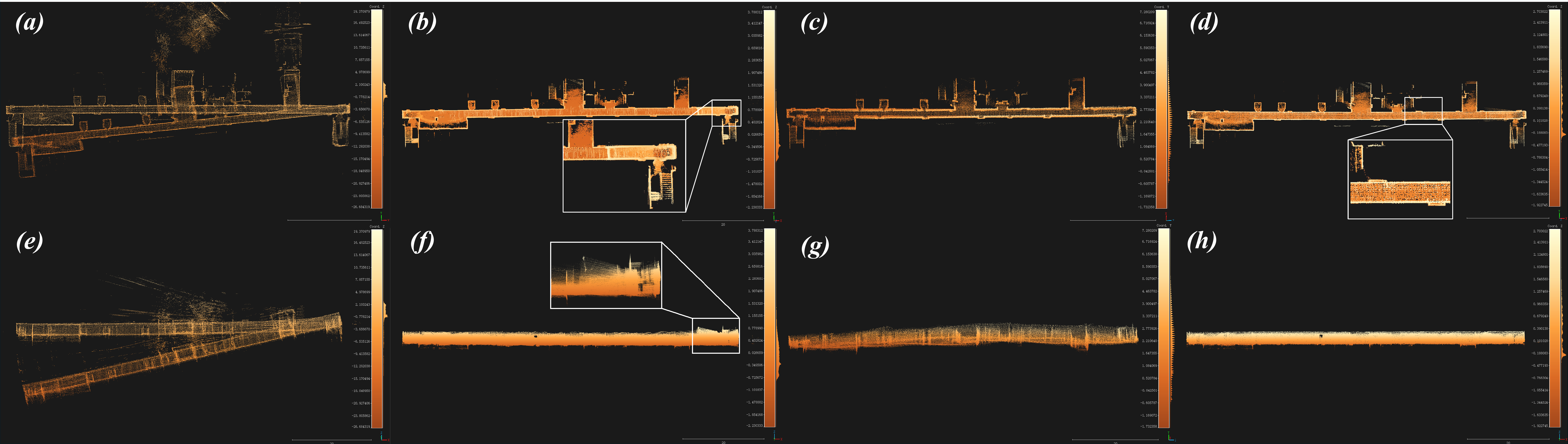}
	\caption{3D point cloud maps constructed by LOAM, FAST-LIO2, LEGO-LOAM, and AIMS in the long corridor experiment.}
	\label{fig:corr_map_all}
\end{figure*}

Fig.~\ref{fig:corr_map_all}(a)-(d) presents the top-view point cloud maps generated by LOAM, FAST-LIO2, LEGO-LOAM, and the proposed method. The color bar indicates height distribution. The environment is flat without slopes. The proposed method produces clear maps with normal wall thickness and no abnormal height points. LOAM exhibits downward degeneration at point $b$, leading to mapping failure. FAST-LIO2 and LEGO-LOAM produce visually similar maps in top view, but abnormal height points can be observed from the color bar, indicating unreliable $z$-axis estimation.

Fig.~\ref{fig:corr_map_all}(e)-(g) shows the side-view maps. FAST-LIO2 exhibits upward motion during the turning maneuver at point $b$. LEGO-LOAM also shows unstable $z$-axis estimation, resulting in backward and downward drifts. The proposed method demonstrates superior robustness.

\subsection{Indoor Garage Corridor Experiment}
To further validate the capability of the proposed method in detecting and handling degenerate scenarios, additional experiments are conducted in an indoor garage laboratory at Chongqing University of Posts and Telecommunications.

In this experiment, the quadrupedal robot starts from the deep end of a corridor inside the garage, moves forward until exiting the corridor into an open area, then rotates clockwise by $90^\circ$, re-enters the corridor, and finally returns to the initial position along the same path. The corridor inside the garage contains very limited geometric features, making LiDAR perception prone to degeneration.

\begin{figure}[t]
	\centering
	\includegraphics[width=\linewidth]{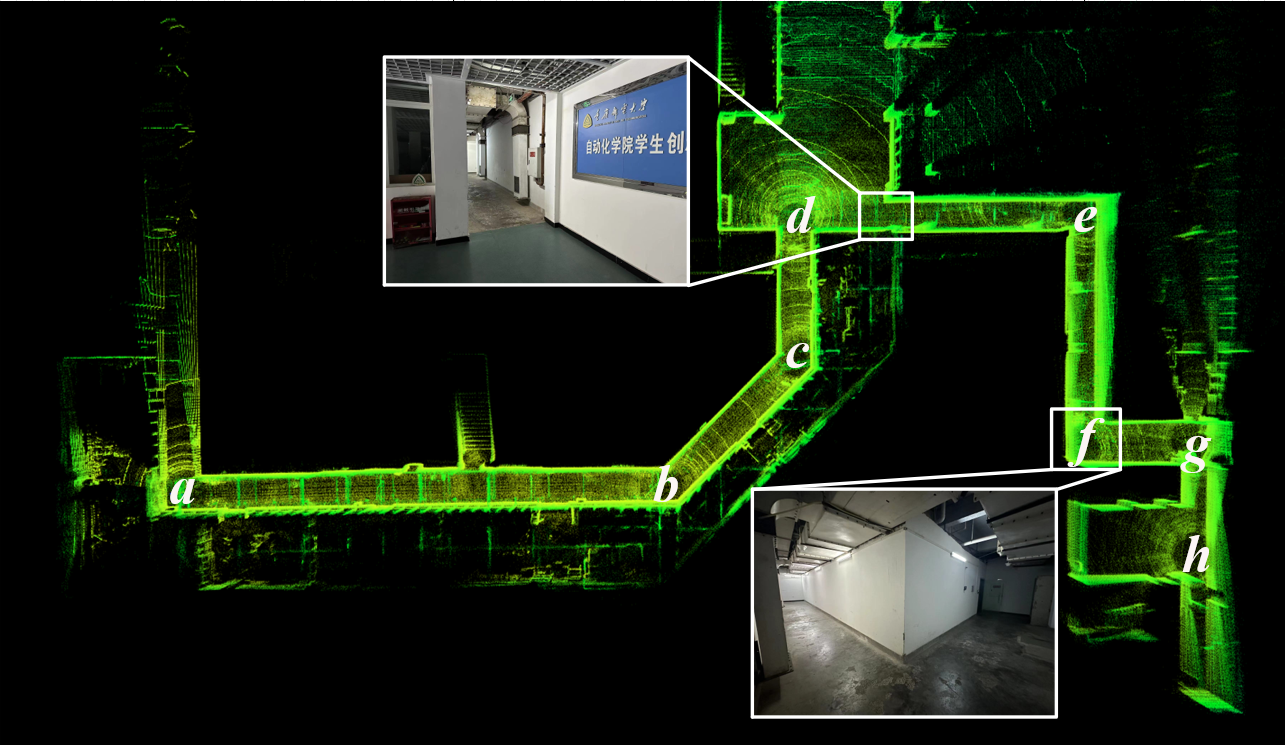}
	\caption{The 3D point cloud map constructed by the proposed method in the garage corridor experiment.}
	\label{fig:proposed_method_map_garage}
\end{figure}

Fig.~\ref{fig:proposed_method_map_garage} illustrates the 3D point cloud map constructed by the proposed method. The embedded images show the corresponding real-world scenes, while the color bar on the right indicates the height distribution of the point cloud. The generated map is clear, with consistent wall thickness and no abnormal height artifacts.

Following the same evaluation procedure as in the previous experiment, eight reference points are selected along the robot trajectory, as illustrated in Fig.~\ref{fig:proposed_method_map_garage}. The path segments from point $a$ to point $d$ and from point $d$ to point $h$ correspond to feature-sparse corridor regions, whereas point $d$ is located in a feature-rich open area. During data collection, the robot moves from point $a$ to point $h$ and then returns to point $a$ along the same path. The straight-line distances between each pair of points are manually measured.

\begin{figure}[t]
	\centering
	\includegraphics[width=\linewidth]{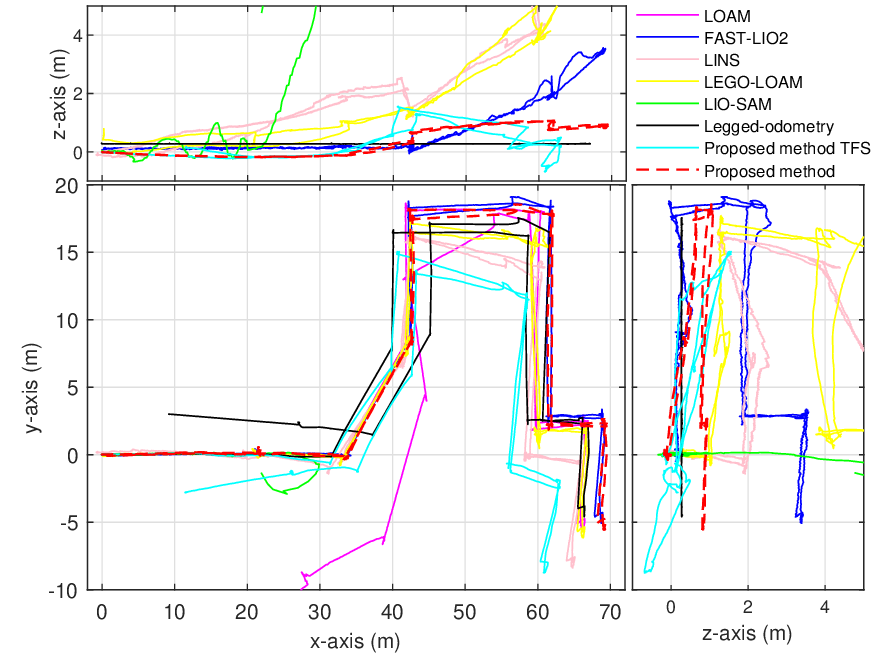}
	\caption{Trajectories estimated by legged odometry and other methods in the garage corridor experiment.}
	\label{fig:all_traj_garage_loog}
\end{figure}

The estimated trajectories are shown in Fig.~\ref{fig:all_traj_garage_loog}. From the trajectory comparison, it can be observed that the leg odometry measurements exhibit significant drift over time. Starting from the origin, FAST-LIO2 and the proposed method follow similar paths. However, after the robot enters the corridor from point $d$, FAST-LIO2 gradually drifts upward due to insufficient ground constraints, which is confirmed by the estimated $z$-axis component.

LINS and LEGO-LOAM suffer from degeneration not only along the forward direction but also along the $z$-axis. Proposed method TFS accumulates larger errors because it fuses erroneous measurements indiscriminately without adaptive weighting. LIO-SAM also experiences early divergence due to the low IMU sampling rate and the extreme perceptual degradation of the environment.

During the experiment, the robot is carefully aligned with the predefined reference points, and the corresponding distances are manually measured to ensure reliable evaluation. Since both the starting and ending positions are located at point $a$, the degeneration error and endpoint drift error are computed.

Table~\ref{tab:garage} reports the quantitative results, where $(a\!-\!d)$ and $(d\!-\!h)$ represent the degeneration degree while the robot traverses the corridor.
\begin{table}[t]
	\centering
	\caption{Quantitative Comparison on the Garage Corridor Dataset}
	\label{tab:garage}
	\begin{tabular}{lccccc}
		\hline
		Method & $(a\!-\!d)$ & $(d\!-\!a)$ & $(d\!-\!h)$ & $(h\!-\!d)$ & $M_{\text{end}}$ \\
		\hline
		\multicolumn{6}{c}{\footnotesize{with "-" and bold number indicates meaningless and best result}} \\
		\hline
		LOAM & 0.26 & 10.57 & 4.85 & 5.13 & 23.84 \\
		FAST-LIO2 & 0.20 & 0.62 & 1.56 & 1.69 & 0.48 \\
		LINS & 4.29 & 4.89 & 8.64 & 8.38 & 0.63 \\
		LEGO-LOAM & 2.50 & 3.64 & 6.68 & 6.27 & 0.85 \\
		LIO-SAM & 76.5 & - & - & - & - \\
		Legged Odometry & 5.87 & 5.92 & 6.45 & 6.92 & 9.70 \\
		Proposed method TFS & 8.31 & 9.65 & 7.44 & 7.65 & 11.92 \\
		Proposed method & \textbf{0.15} & \textbf{0.14} & \textbf{0.48} & \textbf{0.51} & \textbf{0.29} \\
		\hline
	\end{tabular}
\end{table}
Compared with the proposed method, Proposed method TFS yields a significantly larger endpoint drift error of 11.92~m. This is mainly due to two reasons: (1) leg odometry packet loss is not adaptively handled, and (2) degenerate scenarios are not detected, resulting in indiscriminate fusion of unreliable measurements. LOAM exhibits smaller degeneration in the forward path but suffers from large orientation errors during turning maneuvers due to the absence of IMU fusion. LEGO-LOAM and LINS perform reasonably well before entering the corridor but gradually diverge in feature-sparse regions. FAST-LIO2 achieves comparable accuracy in the horizontal plane but shows less robust performance along the $z$-axis.

\section{Conclusion}

In this paper, we propose AIMS, an adaptive multisensor fusion method for robust localization of quadrupedal robots operating in long and narrow inspection environments. The proposed method integrates IMU, LiDAR, and leg odometry within an error-state Kalman filtering framework and introduces an online reliability assessment mechanism to detect perceptual degeneration. Experimental results conducted in tunnel-like environments demonstrate the robustness and effectiveness of AIMS, indicating its practical value for quadrupedal robot inspection applications.

\bibliographystyle{IEEEtran}
\bibliography{references}

\end{document}